\documentclass[11pt,a4paper]{article}
\usepackage[hyperref]{emnlp2020}
\usepackage{times}
\usepackage{latexsym}

\usepackage{multirow}
\usepackage{tabularx}
\usepackage{amsmath}
\usepackage{amssymb}
\usepackage{bm}
\usepackage{subfigure}
\usepackage{graphicx}
\usepackage{latexsym}
\usepackage{arydshln}
\usepackage{bbm}
\usepackage{xpinyin}

\usepackage{microtype}

\aclfinalcopy 
\def\aclpaperid{1770} 

\title{Token-level Adaptive Training for Neural Machine Translation}

\author{Shuhao Gu\textsuperscript{\rm 1,2}, Jinchao Zhang\textsuperscript{\rm 3}, Fandong Meng\textsuperscript{\rm 3}, Yang Feng\textsuperscript{\rm 1,2}\thanks{Corresponding author: Yang Feng. \newline \indent ~~ Joint work with Pattern Recognition Center, WeChat AI, Tencent Inc, China.\newline \indent ~~ Reproducible code: https://github.com/ictnlp/TLAT-NMT. }, \\ \textbf{Wanying Xie\textsuperscript{\rm 1,4}, Jie Zhou\textsuperscript{\rm 3}, Dong Yu\textsuperscript{\rm 4}}\\ 
\textsuperscript{\rm 1} Key Laboratory of Intelligent Information Processing,\\ Institute of Computing Technology, Chinese Academy of Sciences (ICT/CAS)\\
\textsuperscript{\rm 2} University of Chinese Academy of Sciences\\
\textsuperscript{\rm 3} Pattern Recognition Center, WeChat AI, Tencent Inc, China\\
\textsuperscript{\rm 4} Beijing Language and Culture University, China\\
{ \{gushuhao19b,fengyang\}@ict.ac.cn}\\
{ \{dayerzhang,fandongmeng,withtomzhou\}@tencent.com} \\
{xiewanying07@gmail.com, yudong@blcu.edu.cn} \\ 
}

\date{}

\begin{document}
\maketitle

\begin{abstract}
There exists a token imbalance phenomenon in natural language as different tokens appear with different frequencies, which leads to different learning difficulties for tokens in Neural Machine Translation (NMT). 
The vanilla NMT model usually adopts trivial equal-weighted objectives for target tokens with different frequencies and tends to generate more high-frequency tokens and less low-frequency tokens compared with the golden token distribution. 
However, low-frequency tokens may carry critical semantic information that will affect the translation quality once they are neglected.  
In this paper, we explored target token-level adaptive objectives based on token frequencies to assign appropriate weights for each target token during training. 
We aimed that those meaningful but relatively low-frequency words could be assigned with larger weights in objectives to encourage the model to pay more attention to these tokens.
Our method yields consistent improvements in translation quality on ZH-EN, EN-RO, and EN-DE translation tasks, especially on sentences that contain more low-frequency tokens where we can get 1.68, 1.02, and 0.52 BLEU increases compared with baseline, respectively. Further analyses show that our method can also improve the lexical diversity of translation.
\end{abstract}

\begin{table}[t]
    \centering
    \resizebox{1.0\columnwidth}!{
    \begin{tabular}{l|c|c|c}
    \hline
    Token Order &  Average & \multirow{2}{*}{Reference} & Vanilla \\
    (Descending)  & Frequency & ~ & NMT \\
    \hline
    $[0, 10\%)$   &  $10,857$ &$81.75\%$ & $87.26\%$ \\
    $[10\%, 30\%)$ & $516$ &$11.40\%$ & $9.06\%$ \\
    $[30\%, 50\%)$ &$133$ & $3.43\%$ & $2.21\%$ \\
    $[50\%, 70\%)$ &$60$ & $1.95\%$ & $0.99\%$ \\
    $[70\%, 100\%]$ &$25$ &  $1.47\%$ & $0.48\%$ \\
    \hline
    \end{tabular}
    }
    \caption{The average frequency on the NIST training set and proportion of tokens with different frequencies in reference and the translation of the vanilla NMT model (a Transformer model) on the NIST test sets. All the target tokens (BPE sub-words with 30K merge operations ) of the training set are ranked by their frequencies in descending order. The 'Token Order' column represents the frequency interval ($[10\%, 30\%)$ means the frequency of token is between top $10\%$ and $30\%$). The 'Average Frequency' column represents the average frequencies of the tokens in each interval, which show the token imbalance phenomenon in natural language. The last two columns show the vanilla NMT model tends to generate more high-frequency tokens and less low-frequency tokens than reference. }
    \label{tab:exp1}
\end{table}

\section{Introduction}
Neural machine translation (NMT) systems~\cite{kalchbrenner2013recurrent,ChoMGBBSB14,sutskever2014sequence,BahdanauCB14,GehringAGYD17,VaswaniSPUJGKP17} are data driven models, which highly depend on the training corpus. 
NMT models have a tendency towards over-fitting to frequent observations (e.g. words, word co-occurrences) while neglecting those low-frequency observations. 
Unfortunately, there exists a token imbalance phenomenon in natural languages as different tokens appear with different frequencies, which roughly obey the Zipf's Law~\cite{zipf1949human}.  Table~\ref{tab:exp1} shows that there is a serious imbalance between high-frequency tokens and low-frequency tokens. 
NMT models rarely have the opportunity to learn and generate those ground-truth low-frequency tokens in the training process.
Some work tries to improve the rare word translation by maintaining phrase tables or back-off vocabulary~\cite{LuongSLVZ15,JeanCMB15,li2016towards,PhamNW18} or adding extra components~\cite{GulcehreANZB16,zhao2018addressing}, which bring in extra training complexity and computing expense. 
Some NMT techniques which are based on smaller translation granularity can alleviate this issue, such as hybrid word-character-based model~\cite{LuongM16}, BPE-based model~\cite{SennrichHB16a} and word-piece-based model~\cite{wu2016google}. 
These effective work alleviate the token imbalance phenomenon to a certain extent and become the {\em de-facto} standard in most NMT models. 
Although sub-word based NMT models have achieved significant improvements, they still face the token-level frequency imbalance phenomenon, as Table~\ref{tab:exp1} shows. 
Furthermore, current NMT models generally assign equal training weights to target tokens without considering their frequencies. 
It is very likely for NMT models to ignore the loss produced by the low-frequency tokens because of their small proportion in the training sets. The parameters related to them can not be adequately trained, which will, in turn, make NMT models tend to prioritize output fluency over translation adequacy, and ignore the generation of low-frequency tokens during decoding, which is illustrated in Table~\ref{tab:exp1}. It shows that the vanilla NMT model tends to generate more high-frequency tokens and less low-frequency tokens.
However, low-frequency tokens may carry critical semantic information which may affect translation quality once they are neglected.  

To address the above issue, we proposed token-level adaptive training objectives based on target token frequencies. 
We aimed that those meaningful but relatively low-frequency tokens could be assigned with larger loss weights during training so that the model will learn more about them.
To explore suitable adaptive objectives for NMT, we first applied existing adaptive objectives from other tasks to NMT and analyzed their performance. We found that though they could bring modest improvement on the translation of low-frequency tokens, they did much damage to the translation of high-frequency tokens, which led to an obvious degradation on the overall performance. This implies that the objective should ensure the training of high-frequency tokens first.
Then, based on our observations, we proposed two heuristic criteria for designing the token-level adaptive objectives based on the target token frequencies.
Last, we presented two specific forms for different application scenarios according to the criteria.
Our method yields consistent improvements in translation quality on ZH-EN, EN-RO, and EN-DE translation tasks, especially on sentences that contain more low-frequency tokens where we can get 1.68, 1.02, and 0.52 BLEU increases compared with baseline, respectively. Further analyses show that our method can also improve the lexical diversity of translation.

Our contributions can be summarized as follows:
\begin{itemize}
    \item We analyzed the performance of the existing adaptive objectives when they were applied to NMT. Based on our observations, we proposed two heuristic criteria for designing token-level adaptive objectives and present two specific forms to alleviate the problem brought by the token imbalance phenomenon.
    \item The experimental results validate that our method can improve not only the translation quality, especially on those low-frequency tokens, but also the lexical diversity. 
\end{itemize}


\section{Background}
In our work, we apply our method in the framework of \textit{Transformer}~\cite{VaswaniSPUJGKP17} which will be briefly introduced here. 
We denote the input sequence of symbols as $\mathbf{x}=(x_1,\ldots,x_J)$, the ground-truth sequence as $\mathbf{y}^{*}=(y_1^{*},\ldots,y_K^{*})$ and the translation as $\mathbf{y}=(y_1,\ldots,y_K)$.

 \textbf{The Encoder \& Decoder} The encoder is composed of $\mathnormal{N}$ identical layers. Each layer has two sublayers. The first sublayer is a multi-head attention unit used to compute the self-attention of the input, named {\em self-attention multi-head sublayer}, and the second one is a fully connected feed-forward network, named {\em FNN sublayer}.
Both of the sublayers are followed by a residual connection operation and a layer normalization operation. The input sequence $\mathbf{x}$ will be first converted to a sequence of vectors $\mathbf{E}_x=[E_x[x_1];\ldots;E_x[x_J]]$, where $E_x[x_j]$ is the sum of the word embedding and the position embedding of the source word $x_j$. 
Then, this input sequence of vectors will be fed into the encoder and the output of the $\mathnormal{N}$-th layer will be taken as source hidden states. The decoder is also composed of $\mathnormal{N}$ identical layers. In addition to the same kind of two sublayers in each encoder layer, the third {\em cross-attention sublayer} is inserted between them, which performs multi-head attention over the output of the encoder. The final output of the $\mathnormal{N}$-th layer gives the target hidden states $\mathbf{S}=[\mathbf{s}_1;\ldots;\mathbf{s}_I]$, where $\mathbf{s}_i$ is the hidden states of $y_k$. 

 \textbf{The Objective} 
The model is optimized by minimizing a cross-entropy loss with the ground-truth:
\begin{equation}\label{eq::loss}
    \mathcal{L} = -\frac{1}{K} \sum_{k=1}^{K} \log p(y_k^{*} | \mathbf{y}_{<k}, \mathbf{x}),
\end{equation}
where $K$ is the length of the target sentence.

\section{Method}
Our work aims to explore suitable adaptive objectives that can not only improve the learning of low-frequency tokens but also avoid harming the translation quality of high-frequency tokens. 
We first investigated two existing adaptive objectives, which were proposed for solving the token imbalance problems in other tasks, and analyzed their performance. Then, based on our observations, we introduced two heuristic criteria for designing the adaptive objective. Based on the proposed criteria, we put forward two simple but effective functional forms from different perspectives, which can be adapted to various application scenarios in NMT.

\begin{table}[]
\centering
\resizebox{\columnwidth}!{
\begin{tabular}{l|l|ll}
\hline
 & Valid & High & Low \\ \hline
Baseline & 45.46 & 49.27 & 41.35 \\
Linear & 45.33{\em \small{(-0.13)}} & 48.59{\em \small (-0.68)} & 41.64{\em \small (+0.29)} \\ 
Focal & 44.91{\em \small {(-0.55)}} & 48.17{\em \small (-1.10)} & 41.36{\em \small (+0.01)} \\ 
Focal + 1 & 45.71{\em \small {(+0.25)}} & 49.36{\em \small (+0.09)} & 41.93{\em \small (+0.58)} \\\hline
\end{tabular}
}
\caption{BLEU on the validation set of the Chinese-English translation task. 'Low' is the subset of the validation set which contains more low-frequency tokens while 'High' contains more high-frequency tokens. }
\label{tab:valid}
\end{table}

\subsection{Existing Adaptive Objectives Investigation}
The form of adaptive objective is as follows:
\begin{equation}
    \mathcal{L} = -\frac{1}{I}\sum_{i=1}^I w(y_i)\log p(y_i | \mathbf{y}_{<i}, \mathbf{x}),
\end{equation}
where $w(y_i)$ is the weight assigned to the target token $y_i$, which varies as the token frequency changes. Actually, there are some existing adaptive objectives which have been proven effective for other tasks. 
It can help us understand what is necessary for a suitable adaptive objective for NMT if we apply these methods to it.
The first objective we have investigated is the form in Focal loss~\cite{lin2017focal}, which was proposed for solving the label imbalance problem in the object detection task:
\begin{equation}
     w(y_i) = (1-p(y_i))^\gamma. 
\end{equation}
Although it doesn't utilize the frequency information directly, it actually reduces the weights of the high-frequency classes more because they are usually easier to classify with higher prediction probabilities. We set $\gamma$ to $1$ as suggested by their experiments. We noticed that this method greatly reduced the weights of high-frequency tokens, and the variance of weights is large.
The second is the linear weighting function~\cite{JiangRMR19}, which was proposed for the dialogue response generation task:
\begin{equation}
    w(y_i) = - \frac{\mathtt{Count}(y_i)}{\max (\mathtt{Count}(y_k))} + 1, y_k \in \mathrm{V}_t,
\end{equation}
where $\mathtt{Count}(y_k)$ is the frequency of token $y_k$ in the training set and $\mathrm{V}_t$ denotes the target vocabulary. 
Then, the normalized weights $w(y_i)$, which have a mean of $1$, are assigned to the target tokens. We noticed that the weights of high-frequency tokens are only slightly less than $1$, and the variance of weights is small.
We tested these two objectives on the Chinese to English translation task and the results on the validation set are given in Table~\ref{tab:valid}\footnote{The details about the data will be given in the experiment section}. To verify their effects on the high- and low-frequency tokens, we also divided the validation set into two subsets based on the average token frequency of the sentences, the results of which are also given in Table~\ref{tab:valid}.
It shows that although these two methods can bring modest improvement in the translation of the low-frequency tokens, it does much harm to high-frequency tokens, which has a negative impact on the overall performance. 
We noted that both of these two methods reduced the weights of the high-frequency tokens to different degrees, and we argued that when the high-frequency tokens account for a large proportion in NMT corpus, this hinders the normal training of them. To validate our argument, we simply add 1 to the weighting term of focal loss:
\begin{equation}
     w(y_i) = (1-p(y_i))^\gamma + 1.
\end{equation}
The results are also given in Table~\ref{tab:valid} (Row 5), which indicates that this method actually avoids the damage to the high-frequency tokens.
The overall results indicate that it is not robust enough to improve the learning of low-frequency tokens by reducing the weight of high-frequency tokens during the training of NMT. 
Although our goal is to improve the training of low-frequency tokens, we should first ensure the training of high-frequency tokens, and then increase the weights of low-frequency tokens appropriately.
Based on the above findings, we proposed the following criteria.




\subsection{Heuristic Criteria for Token Weighting}\label{sec::guide}
We proposed two heuristic criteria for designing the token-level training weights:

{\bf Minimum Weight Ensurence}. The training weight of any token in the target vocabulary should be equal to or bigger than $1$, which can be described as: 
\begin{equation}
    \forall y_k \in \mathrm{V}_t, w(y_k) \geq 1 
\end{equation}
Although we can force the model to pay more attention to low-frequency tokens by shrinking the weights of high-frequency tokens, the previous analyses have proved that the training performance is more sensitive to the change of high-frequency tokens' weights due to their large proportion in the training set. A relatively small decrease in the weights of high-frequency tokens will prevent the generation probabilities of ground-truth tokens from ascending continually, which may result in an obvious degradation of the overall performance.
Therefore, we ensure that all the token weights are equal to or bigger than $1$ considering the training stability as well as designing convenience.

{\bf Weights Expectation Range Control}. On the condition that the first criterion is satisfied, those high-frequency tokens could have already been well learned without any extra attention. Now, those low-frequency tokens could be assigned with higher weights. 
Meanwhile, we also need to ensure that the weights of low-frequency tokens can't be too large, or it will hurt the training of high-frequency tokens certainly.
Therefore, the expectation of the training weights on the whole training set should be in $[1, 1+\delta]$:
\begin{equation}
    \frac{\sum_{k=1}^{|\mathrm{V}_t|}\mathtt{Count}(y_k)w(y_k)}{\sum_{k=1}^{|\mathrm{V}_t|}\mathtt{Count}(y_k)}=1+\delta, \delta \geq 0,
\end{equation}
where $|\mathrm{V}_t|$ denotes the size of the target vocabulary, $\delta$ is a relatively small number compared with $1$. 
A larger weight expectation means we allocate larger weights to those low-frequency tokens. In contrast, an appropriate weight expectation as defined in this criterion can help improve the overall performance.

The two criteria proposed here are not the only options for NMT, but the adaptive objective satisfying these two criteria can improve not only the translation performance of low-frequency tokens but also the overall performance based on our experimental observations.

\begin{figure}[t!]
    \centering
    \includegraphics[width=0.75\columnwidth]{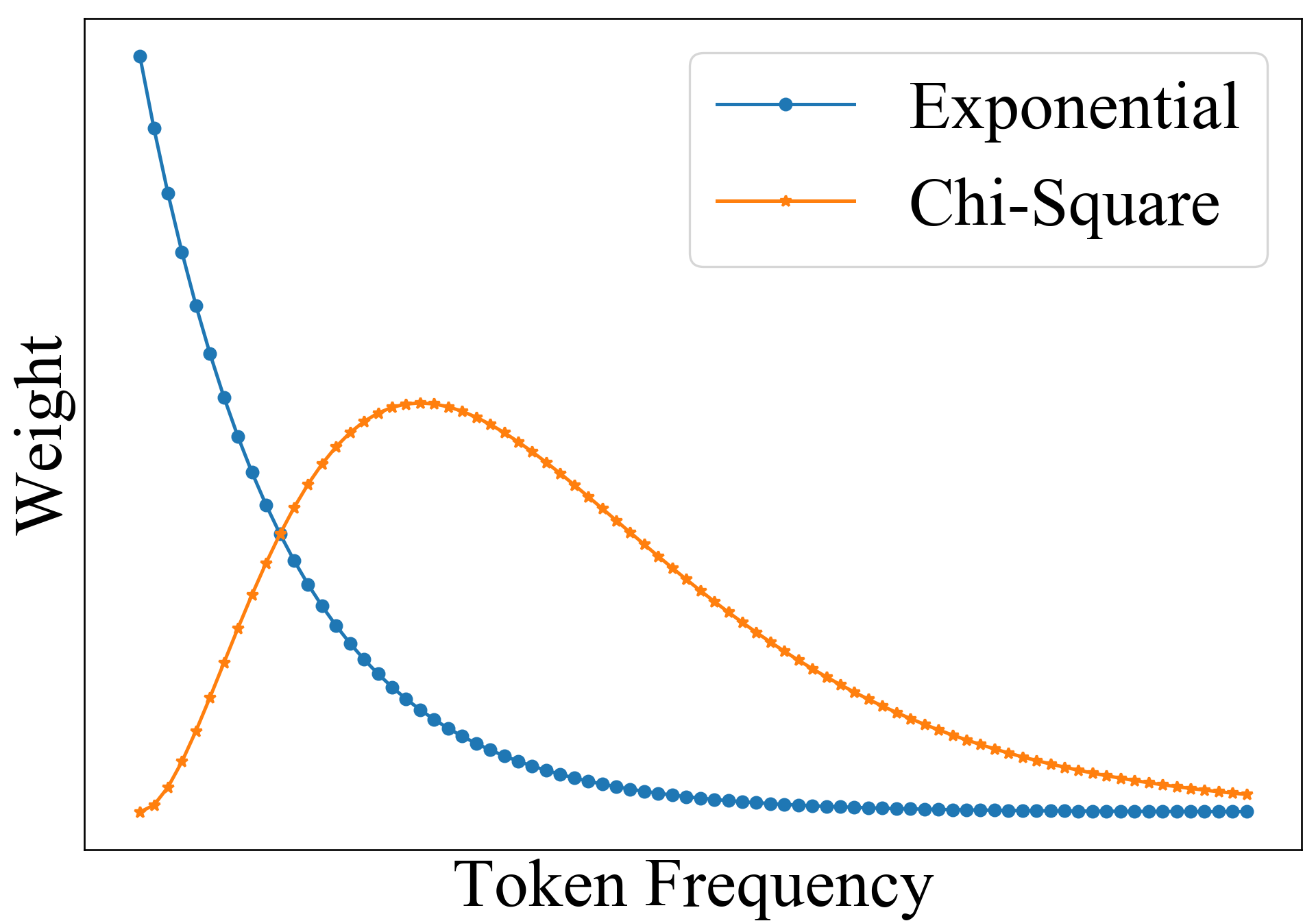}
    \caption{Plots of our two weighting functions. The blue curve is the Exponential form and the orange curve is the Chi-Square form. Both of the hyperparamters are set to $1$.}
    \label{fig:weight}
\end{figure}

\subsection{Two Specific Adaptive Objectives}
In this paper, we proposed two simple functional forms for $w(y_k)$ heuristically based on the previous criteria and justified them with some intuitions. 

\textbf{Exponential:}  
Given the target token $y_k$, we define the exponential weighting function as:
\begin{equation}\label{eq:our_exp}
    w(y_k) = \mathrm{A}\cdot e^{-\mathrm{T}   \cdot \mathtt{Count}(y_k)} + 1.
\end{equation}
There are two hyperparameters in it, i.e., $\mathrm{A}$ and $\mathrm{T}$, which control the shape and the value range of the function. They can be set up according to the two criteria above. The plot of this weighting function is presented in Figure~\ref{fig:weight}. In this case, we don't consider the factor of noisy tokens so that the weight increases monotonically as the frequency decreases. 
Therefore, this weighting function is suitable for cleaner training data where the extremely low-frequency tokens only take up a small proportion. 


\textbf{Chi-Square:} The exponential form weighting function is not suitable for the training data which contain many noisy tokens, because they would be assigned with relatively large weights and have bigger impacts when their weights are summed together. 
To alleviate this problem, we proposed another form of the weighting function:
\begin{equation}\label{eq:our_k2}
    w(y_k) = \mathrm{A} \cdot \mathtt{Count}^2(y_k)e^{-\mathrm{T}\cdot\mathtt{Count}(y_k)} + 1.    
\end{equation}
The form of this function is similar to the form of chi-square distribution, so we named it as chi-square. Plot of this weighting function is presented in Figure~\ref{fig:weight}. We can see from the plot that the weight increases as the frequency decreases at first. Then, after a specific frequency threshold, which is decided by the hyperparameter $\mathrm{T}$, the weight decreases as the frequency decreases. 
In this case, the most frequent tokens and the extremely rare tokens, which could be noise, all will be assigned with small weights. Meanwhile, those middle-frequency words will have larger weights. Most of them are meaningful and valuable for translation but can't be well learned with an equal-weighted objective function. This form of weighting function is suitable for more noisy training data.

\section{Experiments}

\subsection{Data Preparation}

{\bf ZH$\rightarrow$EN}. The training data consists of 1.25M sentence pairs from LDC corpora which has 27.9M Chinese words and 34.5M English words, respectively~\footnote{The corpora include LDC2002E18, LDC2003E07, LDC2003E14, Hansards portion of LDC2004T07, LDC2004T08 and LDC2005T06.}. The data set MT02 was used as validation and MT03, MT04, MT05, MT06, MT08 were used for the test. We tokenized and lowercased English sentences using the Moses scripts\footnote{http://www.statmt.org/moses/}, and segmented the Chinese sentences with the Stanford Segmentor\footnote{https://nlp.stanford.edu/}. The two sides were further segmented into subword units using Byte-Pair Encoding (BPE)~\cite{SennrichHB16a} with $30$K merge operations separately. 

{\bf EN$\rightarrow$RO}. We used the preprocessed version of the WMT2016 English-Romanian dataset released by \citet{lee2018deterministic} which includes 0.6M sentence pairs. We used news-dev 2016 for validation and news-test 2016 for the test. The two languages shared the same vocabulary generated with $40$K merge operations of BPE.

{\bf EN$\rightarrow$DE}. The training data is from WMT2016 which consists of about 4.5M sentences pairs with 118M English words and 111M German words. We chose the news test-2013 for validation and news-test 2014 for the test. $32$K merge operations BPE were performed on both sides jointly.

\subsection{Systems}

We used the open-source toolkit called {\em Fairseq-py}~\cite{edunov2017fairseq} released by Facebook as our Transformer system. 


\noindent \textbullet \ \textbf{Baseline}. The baseline system was implemented as the base model configuration in~\citet{VaswaniSPUJGKP17} strictly. Since our method is further trained based on the pre-trained model at a low learning rate, we also trained another baseline model following the same procedures as our methods have except that all the target tokens share equal weights in the objective, denoted as \textbf{Baseline-FT}. 

\noindent \textbullet \ \textbf{Fine Tuning}~\cite{luong2015stanford}. 
This model was first trained with all the training sentence pairs and then further trained with sentences containing more low-frequency tokens. 
To filter out sentences containing more low-frequency tokens, the method in~\citet{PlataniosSNPM19} was adopted as our judging metric with a small modification:
\begin{equation}\label{eq::rarity}
    d_{\mathrm{rarity}}(\mathbf{y}) \triangleq -\frac{1}{I}\sum_{i=1}^{I}\log \frac{\mathtt{Count}(y_i)}{\sum_{k=1}^{|\mathrm{V}_t|}\mathtt{Count}(y_k)}, 
\end{equation}
where $I$ is the sentence length. 
We added a factor $\frac{1}{I}$ to eliminate the influence of sentence length. All the target sentences were ranked by this metric in ascending order and the bottom one third of the training sentences were chosen as the in-domain data. This method tries to utilize frequency information at the sentence level, while our work uses it at the token level in contrast. 

\noindent \textbullet \ \textbf{Sampler}~\cite{chu2017empirical}. This method oversampled the sentences containing more low-frequency tokens filtered by Eq.~\ref{eq::rarity} three times and then concatenated them with the rest of the training data. Thus the NMT model will be trained with more low-frequency tokens in every epoch.

\noindent \textbullet \ \textbf{Entropy Regularization (ER)}~\cite{PereyraTCKH17}. This method was proposed for solving the overconfidence problem, which adds a confidence penalty term to the original objective:
\begin{equation}
    \mathcal{L}_\mathrm{ER} = L - \alpha \frac{1}{I}\sum_{i=1}^{I}p(y_i|\mathbf{x})\log(p(y_i|\mathbf{x})).
\end{equation}
It is known that token imbalance is one of the causes of overconfidence problem~\cite{JiangR18}, so this method may also alleviate the token imbalance problem. We varied $\alpha$ from $0.05$ to $0.4$ and chose the best one according to the results on the validation sets for different languages. Noting that the label smoothing is applied in the vanilla transformer model which has a similar effect on the output, we removed it from the model when we tested this method. 

\noindent \textbullet \ \textbf{Linear}~\cite{JiangRMR19}. This method was proposed for solving the token imbalance problem in the the dialogue response generation task:
\begin{equation}
    w(y_i) = - \frac{\mathtt{Count}(y_i)}{\max (\mathtt{Count}(y_k))} + 1, y_k \in \mathrm{V}_t.
\end{equation}
Then, the normalized weights, which had a mean of $1$, were applied to the training objective.

\noindent \textbullet \ \textbf{Our\_Exp}. This system was first trained with the normal objective (Equation~\ref{eq::loss}), where all the target tokens have the same training weights. Then the model was further trained with the adaptive objective at a low learning rate. The weights were produced by the Exponential form (Equation~\ref{eq:our_exp}). For computing stability, we used $\frac{\mathtt{Count}(y_k)}{\mathrm{C_{median}}}$ instead of $\mathtt{Count}(y_k)$ in the weighting function, where $\mathrm{C_{median}}$ is the median of the token frequency. 

\noindent \textbullet \ \textbf{Our\_K2}. This system was trained following the same procedure as system Our\_Exp except that the training weights were produced by the Chi-Square form (Equation~\ref{eq:our_k2}). 

The translation quality was evaluated by $4$-gram BLEU~\cite{PapineniRWZ02} with the {\em multi-bleu.pl} script. Besides, we used beam search with a beam size of 4 and a length penalty of 0.6 during the decoding process.

\begin{table}[t!]
    \centering
    \begin{tabular}{l|c|p{1.2cm}<{\centering}|p{1.2cm}<{\centering}|p{1.2cm}<{\centering}}
      ~  & $\mathrm{T}$ & ZH-EN& EN-RO & EN-DE  \\
      \hline
      \textbf{Baseline} & - & 45.49 & 33.60 & 25.45 \\
      \hline
      \multirow{8}{*}{\textbf{Our\_Exp}} & 0.25 & 46.07 & - & -\\
      ~ & 0.35 & \textbf{46.28} & - & - \\
      ~ & 0.50 & 46.19 & 34.10 & - \\
      ~ & 0.75 & 46.13 & 34.11 & - \\
      ~ & 1.00 & 46.01 & 34.24 & 26.02 \\
      ~ & 1.25 & - & \textbf{34.26} & 26.01 \\
      ~ & 1.50 & - & 34.15 & 26.06 \\
      ~ & 1.75 & - & 34.15 & \bf{26.10} \\
      ~ & 2.00 & - & - & 26.03 \\
      \hline
      \multirow{7}{*}{\textbf{Our\_K2}} & 1.50 & 46.14 & - & - \\
      ~ & 1.75 & \textbf{46.24} & - & - \\
      ~ & 2.00 & 46.00 & 34.07 & - \\
      ~ & 2.50 & 45.98 & - & \bf{26.06} \\
      ~ & 3.00 & - & 34.07 & 25.93 \\
      ~ & 4.00 & - & \textbf{34.15} & 25.87 \\
      ~ & 5.00 & - & 34.10 & 25.95 \\
      \hline
    \end{tabular}
    \caption{Performance of our methods on the validation sets for all the three language pairs with different hyperparameters $\mathrm{T}$. Although the best hyperparameter for different languages may be different, it is easy for our method to get a stable improvement.}
    \label{tab:hyper}
\end{table}

\subsection{Hyperparameters}\label{app1}
There are two hyperparameters in our weighting functions, $\mathrm{A}$ and $\mathrm{T}$. In our experiments, we fixed $\mathrm{A}$ to narrow search space and the overall weight range is $[1, e]$. We tuned another hyperparameter $\mathrm{T}$ on the validation data sets under the criteria proposed in section~\ref{sec::guide}. The results are shown in Table~\ref{tab:hyper}. According to the results, the best hyperparameters differed across different language pairs. It is affected by the proportion of low-frequency words and high-frequency words. Generally speaking, when the proportion of low-frequency words gets smaller, the hyperparameter $\mathrm{T}$ should be set smaller too. But it also shows that it is easy for our methods to get a stable improvement over the baseline system following the criteria above. Finally, we used the best hyperparameters as found on the validation data sets for the final evaluation of the test data sets. For example, $\mathrm{T}=0.35$ in the exponential form for ZH$\rightarrow$EN and $\mathrm{T}=4.00$ in the chi-square form for EN$\rightarrow$RO. 

\begin{table*}[ht!]
\centering
\resizebox{2.1\columnwidth}!{
\begin{tabular}{l | l l l l l l l   |   l l   |   l l}
\hline
& \multicolumn{7}{|c|}{ \bf{ZH$\rightarrow$EN} } & \multicolumn{2}{|c|}{\bf{EN$\rightarrow$RO}} & \multicolumn{2}{|c}{\bf{EN$\rightarrow$DE}} \\ 
& \bf \textbf{MT03} & \bf \textsc{MT04} & \bf \textsc{MT05} &  \bf \textsc{MT06}  & \bf \textsc{MT08} & \bf \textsc{AVE}  & \bf $\small{\Delta}$ & \bf \textsc{WMT16} & \bf $\Delta$  & \bf \textsc{WMT16} & \bf $\Delta$ \\
\hline
\bf Baseline  & 44.63 & 45.79 & 44.03 & 43.78 & 35.63 & 42.77 &  & 32.85 &   & 27.15 &   \\
\bf Baseline-FT  & 44.69 & 46.24 & 44.01 & 44.33 & 35.83 & 43.02 &    & 33.15 &  & 27.21 &   \\
\hdashline
\bf Fine Tuning  & 45.06 & 46.30 & 45.30 & 43.61 & 34.68 & 42.99 &  {\em \small{-0.03}}  & 33.28 &  {\em \small{+0.13}} & 26.56 &  {\em \small{-0.65}} \\
\bf Sampler & 44.85 &    46.02 & 44.57 &44.04 & 35.02 & 42.90 & {\em \small{-0.12}} & 32.75 & {\em \small{-0.40}} & - & {\em \small{-}} \\
\bf ER & 44.31 & 46.38 & 45.13 & 44.29 & 35.71 & 43.16 & {\em \small{+0.14}} & 33.21 & {\em \small{+0.06}} & 27.19 & {\em \small{-0.02}} \\
\bf Linear & 44.26 & 46.02 & 43.99 & 44.08 & 34.71 & 42.62 & {\em \small{-0.60}} & 33.35 & {\em \small{+0.20}} & 27.37 & {\em \small{+0.16}} \\
\hdashline
\bf {Our\_Exp} & 45.67** & 47.02** & 45.43** & 44.51 & 36.11 & 43.75 &  {\em \small{+0.73}} & \bf{33.77**} &  \bf{\em \small{+0.62}} & \bf{27.60**} &  \bf{\em \small{+0.39}} \\
\bf {Our\_K2} & \bf{45.87**} & \bf{47.07**} & \bf{45.62**} & \bf{44.72} & \bf{36.20} & \bf{43.90} &  \bf{\em \small{+0.88}} & 33.54* &  {\em \small{+0.49}} & 27.51* &  \em \small{+0.30} \\
\hline
\end{tabular}
}
\caption{BLEU scores on three translation tasks. The column of $\Delta$ shows the improvement compared to Baseline-FT.  **  and * mean the improvements over Baseline-FT is statistically significant \cite{collins2005clause} ($\rho < 0.01$ and $\rho < 0.05$, respectively). 
 The results show that our methods can achieve significant improvements on translation quality.} \label{tab-total}
\end{table*}

\begin{table*}[ht]
\centering
\resizebox{2.0\columnwidth}!{
\begin{tabular}{l | l l l | l l l  }
\hline
& \multicolumn{3}{|c|}{ \bf{ZH$\rightarrow$EN} } & \multicolumn{3}{|c}{\bf{EN$\rightarrow$RO}}  \\ 
& \bf \textsc{High} & \bf \textsc{Middle} & \bf \textsc{Low} &  \bf \textsc{High}  & \bf \textsc{Middle} & \bf \textsc{Low}   \\
\hline
\bf Baseline-FT  & 50.88 & 43.06 & 34.90 & 35.68 & 33.61 & 29.86  \\
\hdashline
\bf Fine Tuning & 49.85(\em \small{-1.03}) & 42.68(\em \small{-0.38}) & 35.85(\em \small{+0.95}) & 35.51(\em \small{-0.17}) & 33.45(\em \small{-0.16}) & 30.56(\em \small{+0.70}) \\
\bf Sampler & 49.77 (\em \small{-1.11}) & 42.63(\em \small{-0.43}) & 35.77(\em \small{+0.87}) & 35.22(\em \small{-0.46}) & 33.07(\em \small{-0.54}) & 30.10(\em \small{+0.42})  \\
\bf ER & 50.59 (\em \small{-0.29}) & 42.82(\em \small{-0.25}) & 35.48(\em \small{+0.58}) & 35.66(\em \small{-0.03}) & 33.25(\em \small{-0.36}) & 30.26(\em \small{+0.41}) \\
\bf Linear & 50.21 (\em \small{-0.67}) & 43.06(\em \small{-0.68}) & 35.19(\em \small{+0.29}) & 35.57(\em \small{-0.11}) & 33.65(\em \small{+0.04}) & 30.35(\em \small{+0.49}) \\
\hdashline
\bf Our\_Exp & 50.88(\em \small{+0.00}) & 43.30(\em \small{+0.24}) & 36.45**(\em \small{+1.55}) & \bf{36.08(\em \small{+0.40})} & \bf{34.26*(\em \small{+0.65})} & \bf{30.88**(\em \small{+1.02})}  \\
\bf Our\_K2 & \bf{51.07(\em \small{+0.19})} & \bf{43.31(\em \small{+0.25})} & \bf{36.58**(\em \small{+1.68})} & 35.94(\em \small{+0.26}) & 33.97(\em \small{+0.36}) & 30.65**(\em \small{+0.79}) \\
\hline
\end{tabular}
}
\caption{BLEU scores on different test subsets which are grouped by their rarities according to Eq.~\ref{eq::rarity}. Sentences in the `Low' contain more low-frequency tokens while the `High' is reverse. The results show that our methods can improve the translation of low-frequency tokens significantly without hurting the translation of high-frequency tokens.} \label{tab-analysis1}
\end{table*}

\begin{table}[t]
\centering
\resizebox{1.0\columnwidth}!{
\begin{tabular}{l | l l l   }
\hline 
& \bf \textsc{High} & \bf \textsc{Middle} & \bf \textsc{Low}   \\
\hline
\bf Baseline-FT  & 28.88 & 26.97 & 25.55   \\
\hdashline
\bf Fine Tuning & 26.40(\em \small{-2.48}) & 26.69(\em \small{-0.28}) & 25.84(\em \small{+0.29})  \\
\bf ER & 28.72(\em \small{-0.16}) & 26.86(\em \small{-0.11}) & 25.74(\em \small{+0.19})  \\
\bf Linear & 28.88(\em \small{+0.00}) & 27.07(\em \small{+0.10}) & 25.70(\em \small{+0.15})  \\
\hdashline
\bf Our\_Exp & \bf{28.91(\em \small{+0.03})} & \bf{27.33*(\em \small{+0.36})} & \bf{26.07**(\em \small{+0.52})}   \\
\bf Our\_K2 & 28.90(\em \small{+0.02}) & 27.28*(\em \small{+0.31}) & 25.99*(\em \small{+0.44}) \\
\hline
\end{tabular}
}
\caption{ EN$\rightarrow$DE BLEU scores on different test subsets. The conclusion is identical to that in Table \ref{tab-analysis1}. } \label{tab-analysis2}
\end{table}

\subsection{Main Results} 

The results are shown in Table~\ref{tab-total}. It shows that the contrast methods can not bring stable improvements over the baseline system. 
They bring excessive damages to the translation of high-frequency tokens which can be proved by the analyzing experiments in the next section. 
As a contrast, our methods can bring stable improvements over Baseline-FT almost without any additional computing or storage expense. On the EN$\rightarrow$RO and EN$\rightarrow$DE translation tasks, Our\_Exp is more effective than Our\_K2 while on the ZH$\rightarrow$EN translation task the result is reversed. The reason is that the NIST training data set contains more noisy tokens, which can be ignored by the Our\_K2 method. More analyses based on the token frequency are shown in the next section.

\section{Analysis}
\subsection{Effects on Translation Quality with Considering Token Frequencies}
To further illustrate the effects of our method, we evaluated the performance based on the token frequency. For the ZH$\rightarrow$EN translation task, we concatenated the MT03-08 test sets together as a big test set. For the EN$\rightarrow$RO and EN$\rightarrow$DE translation tasks, we just used their test sets. Each sentence was scored according to Eq.~\ref{eq::rarity} and sorted in ascending order. Then the test set was divided into three subsets with equal size, denoted as \textsc{High}, \textsc{Middle}, and \textsc{Low}, respectively. Sentences in the subset \textsc{Low} contain more low-frequency tokens while the \textsc{High} is reverse.

The results are given in Table~\ref{tab-analysis1} and Table~\ref{tab-analysis2}. 
The contrast methods outperform the Baseline-FT on the \textsc{Low} subset but are worse than it in the \textsc{High} and \textsc{Middle} subsets, which indicates that the gains on the translation of low-frequency tokens come at the expense of the translation of high-frequency tokens.
As a contrast, both of our methods can not only bring a significant improvement on the \textsc{Low} subset but also get a modest improvement on the \textsc{High} and \textsc{Middle} subsets.
It can be concluded that our methods can ameliorate the translation of low-frequency tokens without hurting the translation of high-frequency tokens.

\begin{figure}[t]
    \centering
    \includegraphics[width=0.8\columnwidth]{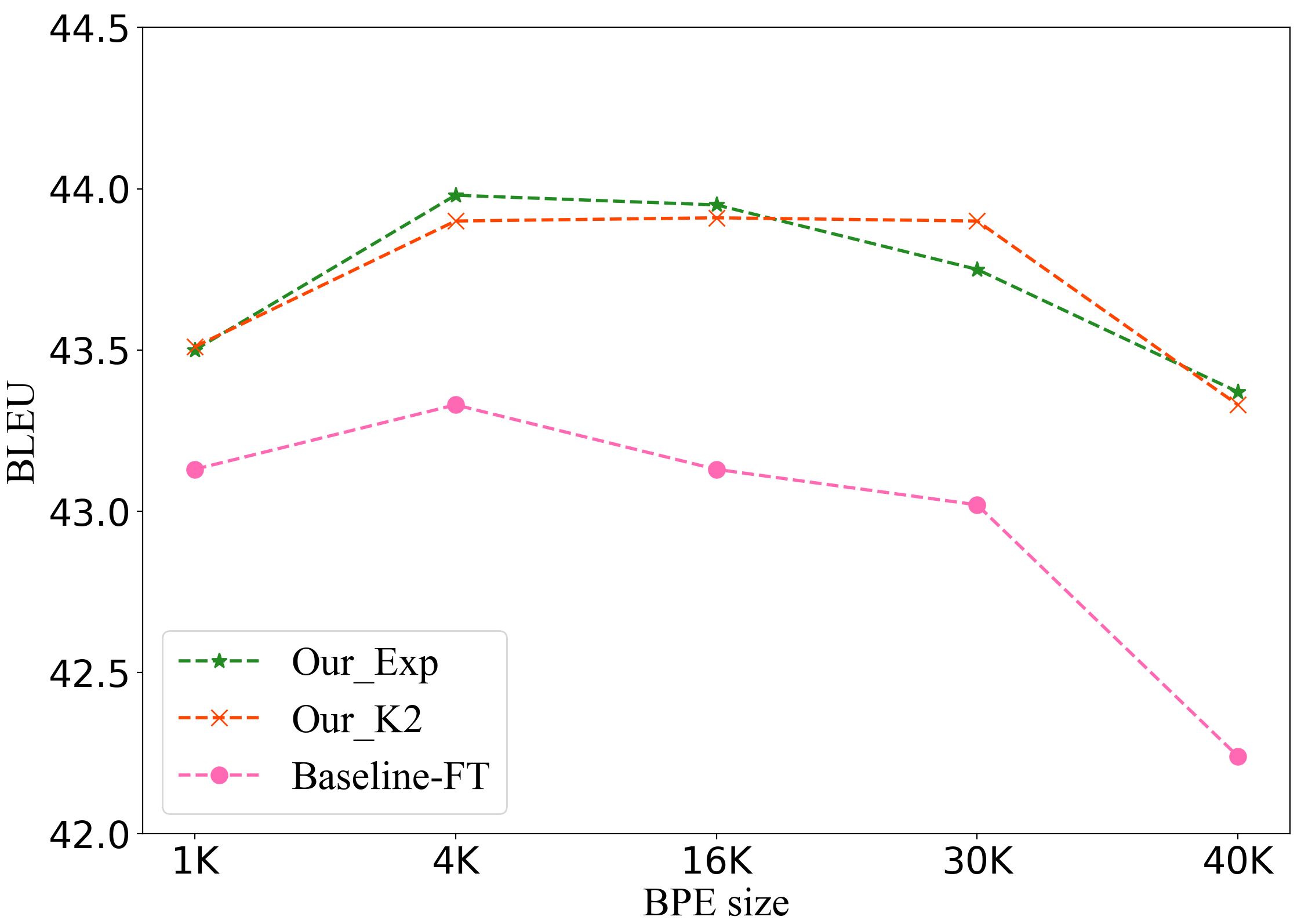}
    \caption{BLEU with different BPE sizes on ZH$\rightarrow$EN translation task. It shows that our method can always bring a stable improvement compared with the baseline.}
    \label{fig:bpe}
\end{figure}

\subsection{Effects on Translation Quality with Different BPE Sizes}
It is known that the BPE sizes have a large impact on the data distribution. Intuitively, a smaller size of BPE will bring a more balanced data distribution, but it will also increase the average sentence length and neglect some token co-occurrences. To verify the effectiveness of our method with different BPE sizes, we varied the BPE sizes from $1$K to $40$K on the ZH$\rightarrow$EN translation task. The results are shown in Figure~\ref{fig:bpe}. It shows that as the number of BPE size increases, the BLEU of baseline rises first and then declines. Compared with the baseline systems, our method can always bring improvements, and the larger the BPE size, i.e., the more imbalanced the data distribution, the larger the improvement brought by our method. In practice, the BPE size either comes from the experience or is chosen from several trial-and-errors. No matter what the situation is, our method can always bring a stable improvement. 

\begin{figure}[t]
    \centering
    \includegraphics[width=0.85\columnwidth]{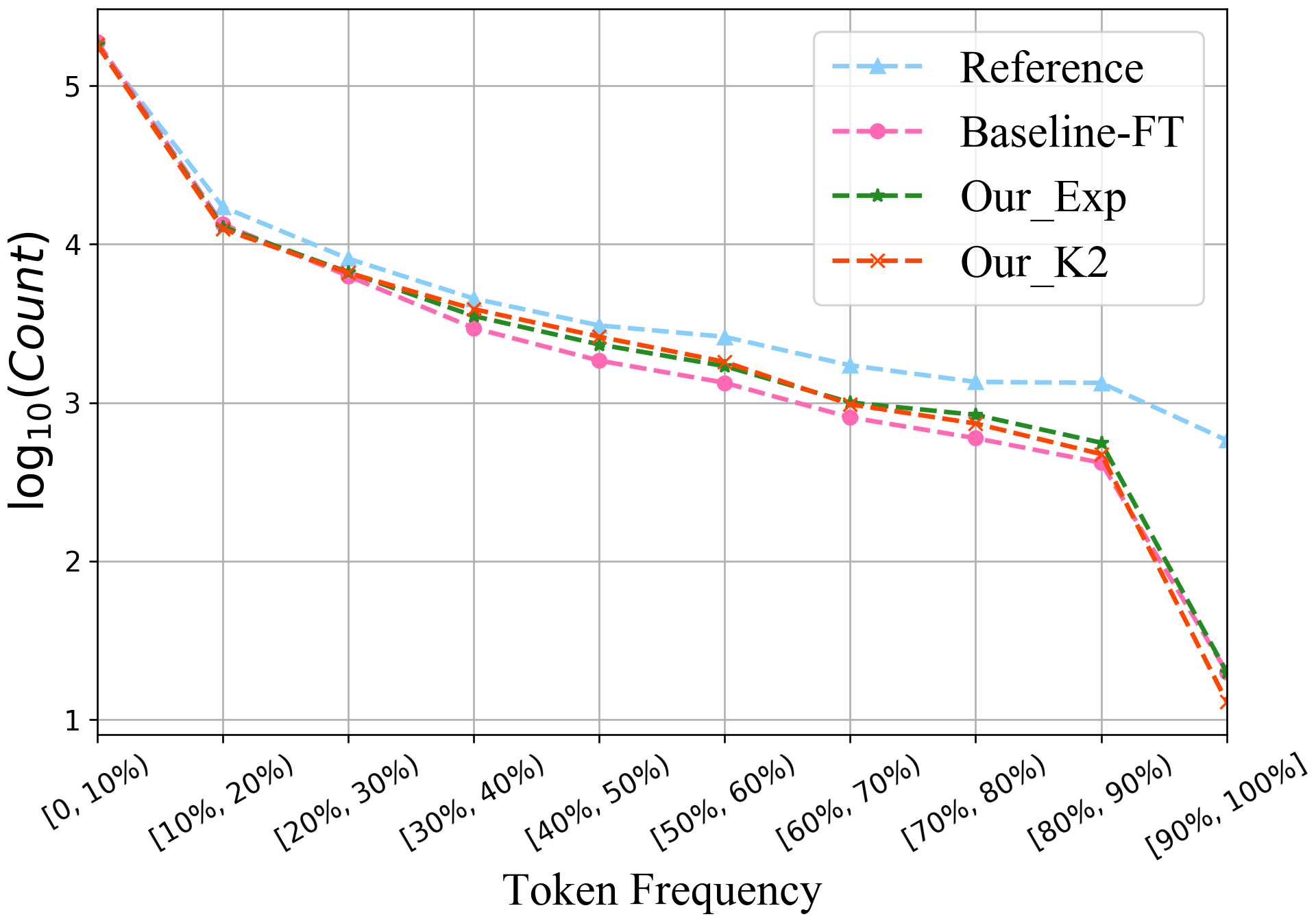}
    \caption{The count of tokens with different frequencies in references, translations of the baseline systems and our methods on the ZH$\rightarrow$EN translation task. The tokens are ranked by their frequencies in the training sets. The x-axis represents the frequency interval ([20\%, 30\%) means the frequency of tokes is between top 20\% and 30\%), the y-axis is the count of the tokens applied with a common logarithm operation in each interval. }
    \label{fig:1}
\end{figure}

\begin{table}[t]
\centering
\resizebox{\columnwidth}!{
\begin{tabular}{l|ccc}
\hline
  & TTR\small{($\times 10^{-2}$)} & HD-D & MTLD \\ \hline \hline
Baseline-FT & 5.32 & 0.829 &59.1  \\
Our\_Exp & 5.87 & 0.836 & 62.2 \\ 
Our\_K2 & 5.95 & 0.835 & 61.9  \\ 
Reference & 6.79 & 0.852 & 69.2  \\
\hline
\end{tabular}
}
\caption{The lexical diversity of translations. A larger value represents higher diversity. The results show that our method can improve the lexical diversity.}
\label{tab:diversity}
\end{table}

\subsection{Effects on Token Distribution and Lexical Diversity}
Compared with the reference, the outputs of the vanilla NMT model contain more high-frequency tokens and have lower lexical diversity~\cite{VanmassenhoveSW19}.   
To verify whether our methods can alleviate these problems, we did the following experiments based on the ZH$\rightarrow$EN translation task. 
The tokens in the target vocabulary were first arranged in descending order according to their token frequencies. Then they were divided into ten intervals equally. Finally, we counted the number of tokens in each token frequency interval of the reference and the translation of different systems. The results are shown in Figure~\ref{fig:1} and we did a common logarithm for display convenience. It shows that there is an obvious gap between the Baseline-FT and reference, and the curve of Baseline-FT is lower than the curve of reference in every frequency interval except for the top 10\%. As a contrast, our methods can reduce this gap, and the tokens distribution is closer to the real distribution. 
Besides, we also measure the lexical diversity of the translations with several criteria, namely, type-token ratio (TTR)~\cite{herrick1958certain}, the approximation of hypergeometric distribution (HD-D) and the measure of textual lexical diversity (MTLD)~\cite{mccarthy2010mtld}. The results are given in Table~\ref{tab:diversity}. It shows that our method can also improve the lexical diversity of the translation.

\begin{table}[t]
    \centering
    \resizebox{1.05\columnwidth}!{
    \begin{tabular}{l|l}
    \multirow{2}{*}{\bf{ Source}}     &  búduàn  guānbì  nàxiē  wūrǎn  huánjìng  \\ 
    ~ & de  \textbf{méikuàng} . \\
       \hline
       \multirow{2}{*}{\textbf{Reference}}   & those \textbf{coalmines} pollute the environment  \\ 
       ~ & should be continuously shut down . \\
       \hline
        \multirow{2}{*}{\textbf{Baselie-FT}}   & continually close down those {\bf mines} \\ 
        ~ & that pollute the environment . \\
        \hline
       \multirow{2}{*}{\textbf{Our\_Exp}}   & those {\bf coalmines} that pollute the environment  \\ 
       ~ & should be continuously closed. \\
        \hline
       \multirow{2}{*}{\textbf{Our\_K2}}   & those {\bf coalmines} that pollute the environment \\ 
       ~ & should be continuously closed. \\
        \hline
        \hline
        \textbf{Source} & yǐhòu kěyǐ gěi wǒ  dāndú \textbf{pèi} jiān bàngōngshì . \\
        \hline
        \textbf{Reference} & an exclusive office could be \textbf{assigned} me later on . \\
        \hline
        \textbf{Baselie-FT} & later i could \textbf{match} my office alone . \\
        \hline
        \textbf{Our\_Exp} &  i could be \textbf{assigned} an office alone later . \\
        \hline
        \textbf{Our\_K2} &  later i could be \textbf{assigned} an office alone . \\
        \hline

    \end{tabular}
    }
    \caption{Translation examples of the Basline-FT and our methods. The results show that our methods can generate low-frequency but more accurate tokens.}
    \label{tab:case}
\end{table}

\subsection{Case Study}
Table~\ref{tab:case} shows two translation examples in the ZH$\rightarrow$EN translation direction. In the first sentence, the Baseline-FT system failed to generate the low-frequency noun `{\em coalmine}' (frequency: $43$), but generated a relatively high-frequency word `{\em mine}' (frequency: $1155$). We can see that this low-frequency token carries the central information of this sentence, and the mistranslation of it prevents people from understanding this sentence correctly. In the second sentence, our methods generated the low-frequency verb `{\em assigned}' (frequency: $841$) correctly, while the Baseline-FT generated a more frequent token `{\em match}' (frequency: 1933), which reduced the translation accuracy and fluency. These examples can be part of the evidence to show the effectiveness of our methods.

\section{Related Work}
\textbf{Rare Word Translation}. Rare word translation is one of the key challenges for NMT. For word-level NMT models, NMT has its limitation in handling a larger vocabulary because of the training complexity and computing expense. 
Some work tries to solve this problem by maintaining phrase tables or back-off vocabulary~\cite{LuongSLVZ15,JeanCMB15,li2016towards}.
The subword-based NMT~\cite{SennrichHB16a,LuongM16,wu2016google} reduces the size of vocabulary greatly and become the mainstream technology gradually. \citet{abs-2004-02334} gave a detailed analysis about the effects of the BPE size on the data distribution and translation quality. Some recent work tried to further improve the translation of the rare words with the help of the memory network or the pointer network~\cite{zhao2018addressing,PhamNW18}. 
In contrast, our methods can improve the translation performance without extra cost and can be combined with other techniques. 

\textbf{Class Imbalance}. Class imbalance means the total number of some classes of data is far less than the total number of other classes. This problem can be observed in various tasks~\cite{wei2013effective,johnson2019survey}. 
In NMT, the class imbalance problem might be the underlying cause of, among others, the gender-biased output problem~\cite{abs-1909-05088}, the inability of MT system to handle morphologically richer language correctly~\cite{PassbanWL18}, or the exposure bias problem~\cite{RanzatoCAZ15,ShaoCF18,ZhangFMYL19}.
The methods of trying to solve this can be divided into two types. The data-based methods~\cite{baloch2015investigation,ofek2017fast} make use of over- and under-sampling to reduce the imbalance. The algorithm-based methods~\cite{zhou2005training,lin2017focal} give extra reward to different classes. 
Our method is algorithm-based which brings no extra cost. 

\textbf{Word Frequency-based Methods}. Some work also makes use of word frequency information to help learning, such as in the word segmentation~\cite{sun2014feature} and term extraction~\cite{frantzi1998c,vu2008term}. In NMT, word frequency information is used for curriculum learning~\cite{KocmiB17,abs-1811-00739,PlataniosSNPM19} and domain adaptation data selection~\cite{WangULCS17,zhang2018sentence,GuFL19}. \citet{WangTSL20} analyzed the miscalibration problem on the low-frequency tokens.
\citet{JiangRMR19} proposed a linear weighting function to solve the word imbalance problem in the dialogue response generation task. Compared with it, our method is more suitable for NMT.  

\section{Conclusion}
In this work, we focus on the token imbalance problem of NMT. We show that the output of vanilla NMT contains more high-frequency tokens and has lower lexical diversity.
To alleviate this problem, we investigated existing adaptive objectives for other tasks and then proposed two heuristic criteria based on the observations. Next, we gave two simple but effective forms based on the criteria, which can assign appropriate training weights to target tokens.
The final results show that our methods can achieve significant improvement in performance, especially on sentences that contain more low-frequency tokens. Further analyses show that our method can also improve the lexical diversity.

\section*{Acknowledgements}
We thank all the anonymous reviewers for their insightful and valuable comments. This work was supported by National Key R\&D Program of China (NO. 2017YFE0192900).

\bibliography{emnlp2020}
\bibliographystyle{acl_natbib}


\end{document}